\documentclass[graybox]{svmult}
\pdfoutput=1

\usepackage{mathptmx}       
\usepackage{helvet}         
\usepackage{courier}        
\usepackage{type1cm}        
\usepackage{makeidx}         
\usepackage{graphicx}        
\usepackage{multicol}        
\usepackage[bottom]{footmisc}

\usepackage{makecell}
\usepackage{natbib}

\begin{document}

\title*{Features selection in NBA outcome prediction through Deep Learning}
\titlerunning{Features selection in NBA outcome prediction}
\author{Manlio Migliorati}
\institute{Manlio Migliorati \at
              University of Brescia, Italy\\
              \email{manlio.migliorati@unibs.it}          
}
%
%
\maketitle

\abstract{This manuscript is focused on features' definition for the outcome prediction of matches of NBA basketball championship. It is shown how models based on one a single feature (Elo rating or the relative victory frequency) have a quality of fit better than models using box-score predictors (e.g. the Four Factors). Features have been ex ante calculated for a dataset containing data of 16 NBA regular seasons, paying particular attention to home court factor. Models have been produced via Deep Learning, using cross validation.}

\keywords{basketball outcome prediction, features definition, court factor}

\section{Introduction}
\label{intro}
This manuscript is focused on features definition for the problem of predicting the winner in NBA matches. It is shown how, for this classification problem, a careful definition of single features to be used in machine learning techniques, and Deep Learning \citep{Goodfellow2016, Chollet2018} in particular, can produce predictions with a quality better than quality of models built on the top of \emph{box-score} statistics as, for instance, Oliver's Four Factors \citep{Oliver2004,Kubatko2007}.
\\To this purpose, two features directly quantifying strength of teams involved in a match have been selected:
\begin{enumerate}
\item The Elo (from the name of its creator) rating system \citep{Elo1978}, originally defined for rating chess players and today widely used in several domains. 
\item The difference of the relative frequency of victories for the two teams involved in a match (see for instance \citep{Migliorati2020cms} where it is named \(diff\)).
\end{enumerate}
and used as covariates to build models to be compared, in terms of quality of fit, to models built using Oliver's Four Factors \citep{Oliver2004, Kubatko2007}, the famous set of statistics  synthesizing \emph{box-score} information and considered fundamental for winning a match.
\\The dataset includes all the NBA regular seasons matches from 2004-2005 to 2019-2020 (until 11/03/2020, when NBA was stopped for some weeks due to Covid19). For each observation, features' values have been calculated ex-ante, i.e. considering only information about prior matches, and taking into account:
\begin{itemize}
\item both historical (averaging considering all past games) and dynamic (averaging considering only some prior matches) features' definitions.
\item Regression to mean \citep{Galton1889}, i.e. the extreme values trend of getting closer to average in repeating experiments; this concept is particularly important in NBA championship, where after each season's end there is an explicit attempt of strength re-balancing among teams (draft mechanism).
\item The court factor, so important in NBA \citep{Harville1994, Jones2007}. Beside classical features calculation, considering all past matches regardless the court where they have been played, for each feature two new supplementary statistics have been calculated, taking itno account only information either from home or from away matches, respectively. In this way we will have not only values for Elo, \(diff\) and Four Factors regressors, but also for home Elo and away Elo, for home \(diff\) and away \(diff\), and for home Four Factors and away Four Factors.
\end{itemize}
Deep Learning is a specific subfield of machine learning, and is based on using neural networks composed by (possibly) several layers (this is the reason why the approach is called deep, no relevance to a deeper data understanding level).
\\The models used in this work have been developed in a particular Deep Learning echosystem in R, using {\fontfamily{pcr}\selectfont Keras} package \citep{Allaire2021} via RStudio \citep{RStudioTeam2021}. The nets, calibrated to produce models with a good prediction quality, are built considering the two hyperparameters (i.e. the number of layers and the numbers of units for each layer) small in size, a natural consequence of the actually restricted number of features.
This manuscript's original contributions are related to:
\begin{enumerate}
\item the comparison of single-feature and box-score based models, showing how the latter have a lower prediction quality (a possible symptom of the fact that they are close to their limit in outcome prediction).
\item the usage of really simple neural networks for a complex classification problem, asking for a reduced computational power and producing results comparable, for the specific problem, to results produced by more complex network architectures.
\item the building of two new variants for the features we are considering. These variants are calculated considering either only data from the home played matches or only data from the away matches, respectively.
\end{enumerate}
The manuscript is organized as follows: Section \ref{s:d} contains the review of literature about basketball predictions via machine learning, Section \ref{s:fd} contains the formalization of the definition for the selected features, Section \ref{s:dataset} is devoted to describe the dataset and how features have been characterized; Section \ref{sec:ann:mm} summarizes Deep Learning approach; Section \ref{sec:ann:res} reports the outcome prediction results produced in applying Deep Learning, and Section \ref{sec:ann:con} proposes some conclusions and ideas for future enhancements.

\section{Literature review: basketball predictions via machine learning}
\label{s:d}

It is several years that data analytics play a fundamental role in sport analysis and management: in the last decades, publications on statistics in sport have multiplied, and a data-based approach was adopted in each professional sport \citep{Alamar2013,Albert2017}, facing different kinds of problems. \\Analysis and applications of statistics to sport include performance measurement \citep{Mackenzie2013, Page2007, Passos2016, Sandri2020, Zuccolotto2017a,  Zuccolotto2019}, injuries prevention (see \citealp{VanEetvelde2021} for a review), optimal game strategies \citep{Zuccolotto2020}, match preparation \citep{Migliorati2020, Miller2015, Thabtah2019}, players' selection \citep{Lewis2003} and, of course, outcomes forecasting \citep{Bunker2019, Wunderlich2020}.
\\In effect, it was with the application of the data-driven approach described in \citealp{Lewis2003}, centered on selection of players for Oakland Athletics baseball team, that analytics in sport actually entered the maturity phase.
\\Then, quickly, data mining in sport has been widely adopted and adapted in all professional sports, such as baseball (maybe the sport with the greatest history in analytics, starting in 1977 with dedicated reports) \citep{Koseler2017}, hockey (see \citealp{Swartz2017} for a review), American football \citep{ Baker2015, Silver2014}, football \citep{Carpita2015, Carpita2020, Sarmento2014} and, of course, basketball.
\\Basketball milestones of this analytics-based approach are pioneering works \citep{Kubatko2007, Oliver2004}, where the famous Oliver's ``Four Factors to success'' were introduced as four indexes containing a big amount of information. Then, a huge number of analyses have been done applying data mining to basketball data (see, for example,\citealp{Bianchi2017, Groll2018, Metulini2018, Sandri2020, Zuccolotto2017a, Zuccolotto2017b, Zuccolotto2019, Zuccolotto2020}).

Considering the large interest and the increasing volume in sport betting, it is easy to understand the reason why the number of attempts in predicting games' results is continuously increasing, see for instance \citealp{Bunker2019, Hubacek2019}.
\\Machine learning techniques for outcome prediction have been widely applied \citep{Haghighat2013}, covering all professional sports, from horse races \citep{Davoodi2010} to hockey \citep{Gu2016} and from American football \citep{Beal2020, David2011, Kahn2003, Purucker1996} to football \citep{Carpita2019, Min2008, Tax2015}, just to give some examples among others. Also basketball, of course, has been investigated under this perspective.
\\In \citet{Loeffelholz2009} authors worked on a dataset of 650 NBA games, and used several kinds of ANN (Artificial Neural Networks, \citealp{zhang2000}) for outcomes prediction, correctly predicting the winning team 74.33 percent of the time (on average), higher than experts percentage claimed to be 68.67.
\\In \citet{Miljkovic2010} it is reported how, among several machine learning algorithms, best results in both predicting the outcomes and calculating the final match spread were produced by the Naïve Bayes approach. Authors used 778 NBA games of season 2009-2010, considering 141 features as input, and an accuracy of 67\% is reported.
\\In \citet{Cao2012} data of 5 NBA seasons were analyzed using ANN, Support Vector Machine \citep{Cortes1995}, Naïve Bayes and logistic regression, with the latter approach producing the best prediction accuracy (about 70\%) for the classification problem of predicting the winner of a game.
\\In a similar way, in \citet{Beckler2013} authors used Linear Regression, Support Vector Machines, Logistic Regression and ANN for NBA outcomes' prediction, using a dataset including seasons from 1991-1992 to 1996-1997 and reporting an accuracy of 73\%.
\\In \citet{Cheng2016} authors applied the principle of Maximum Entropy \citep{Jaynes1957} to predict NBA playoff outcomes for seasons from 2007–08 to 2014–15, using box score information as features, reporting an accuracy of 74.4\%.
\\At last, there are several betting sites suggesting NBA outcome predictions. As an example, \citet{Teamranking2021} proposes predictions about NBA match winners using 4 approaches, built on the basis of several sources (historical data, breaking news and trends). For regular season 2017-2018 the maximum accuracy is 74.3\%, obtained using decision trees on data of games of March.
\\Many works and many results, with a great difficulty in comparing outcomes predictions related to so different dataset. In this work we prepared just one dataset, and calculate on it all the features to be used  for our goal: to show how models using single features have a quality of fit greater than quality of models based on box-score in general, and Four Factors in particular, where a huge number of independent variables is often used.

\section{Features' definition}
\label{s:fd}

\subsection{The Elo rating}
\label{sss:ann:fd:elo}
\begin{figure}
 \includegraphics[width=0.7\textwidth]{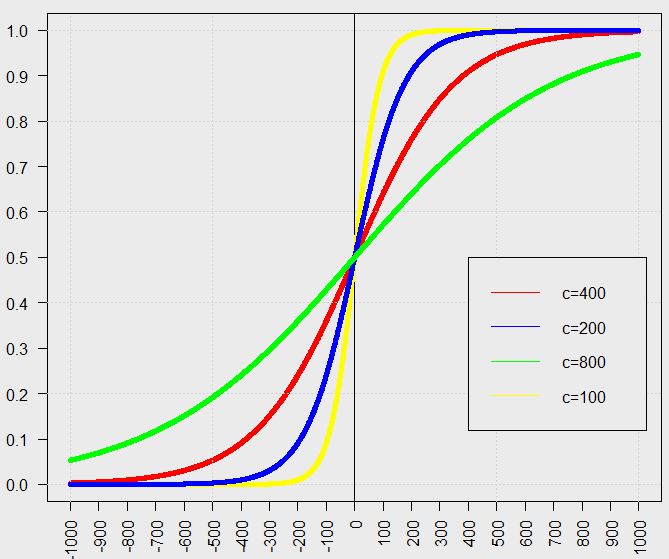}
\caption[Elo logistic curves]{Elo logistic curves on the basis of different logistic parameter value (i.e. the denominator of the exponent in equations \ref{elo.eq1} and \ref{elo.eq2}); 400 is the default in chess.}
 \label{fig.elo} 
\end{figure}

The Elo rating system \citep{Elo1978} has been originally defined for calculating the strength of players in zero-sum games (i.e. games where a player gains exactly what its opponent loses) as chess, the sport for which this system was created by Arpad Elo. The Elo rating system, with some adjustments, has been applied to many sports, mainly for tournament rating predictions: football \citep{Eetvelde2019, Hvattum2010, Leitner2010, WFER2021}, tennis \citep{Angelini2021}, Australian football \citep{Ryall2010}, ice hockey \citep{WFER2021}, American football \citep{Silver2014} and, of course, to NBA basketball \citep{Silver2015,Silver2020}.
\\ Each player is assigned a rating constituted by a single number: new players have an initial default rating (that can change on the basis of the considered organization), and the difference in the ratings of the opponents of a match is used to establish the probability of the match result. After every match, the winning player will gain a certain quantity of points (and the defeated  player will lose the same quantity) depending on their pre-match rating difference; moreover, the system is built in an ``asymmetric'' way: the gain for victory of the player with the highest rating is smaller than the eventual gain for victory of the player with the lowest rating.
\\More formally: if before a match Player1 has a rating \(R1\) and Player 2 has a rating \(R2\), the probability for Player 1 of winning the match  (event \(p1w\)) is modeled as a logistic curve as follows:
\begin{equation}\label{elo.eq1}
P(p1w)=\frac{1}{1+10^{\frac{-(R1-R2)}{400}}}
\end{equation}
and the probability of victory for Player 2 (event \(p2w\)) is modeled as:
\begin{equation}\label{elo.eq2}
P(p2w)=\frac{1}{1+10^{\frac{-(R2-R1)}{400}}}
\end{equation}
\\ where the value 400 is historically used in Elo (Fig. \ref{fig.elo} shows the impact of that parameter on the slope of the sigmoid curve). Probabilities are 0.5 if Player1 and Player2 share the same rating before the match, and in general \(P(p1w)+P(p2w)=1\).

Let \(S\) be the result of a match:  for games without the possibility of draws (as basketball is) \(S\) is 1 for Player1 victory 0  for Player2 victory\footnote{For sports as football, where a draw is admitted too, \(S\) can assume the three possible values 1, 0.5, 0.}.
 After the match, the ratings of the 2 players will be updated as follows:
\begin{equation}
\label{elo.eq3}
 R1\textsuperscript'=R1+K*(S-P(p1w))
 \end{equation}
\begin{equation}
\label{elo.eq4}
 R2\textsuperscript'=R2+K*(S-P(p2w))
\end{equation}
where \(K\) is a parameter addressing how strongly a result will affect ratings' update: a small \(K\) means that ratings remain almost stable, a high \(K\) means strong impacts on rating change.

An example: if the ex-ante ratings are 1500 for Player1 and 1400 for Player2, and Player1 won the match, the updated ratings will be:
\begin{itemize}
\item for \(K\)=5, Player1 =1502 and Player 2=1398
\item for \(K\)=50, Player1 = 1518 and Player2 =1382
\end{itemize}
Viceversa, if Player2  (i.e. the underdog) won the match, larger variations in ratings will be produced:
\begin{itemize}
\item for \(K=5\), Player1 =1497 and Player 2=1403
\item for \(K=50\), Player1 = 1468 and Player2 =1432
\end{itemize}

In the chess world the logistic parameter is set equal to 400, to have P(p1w)= 0.75 and P(p2w) =0.25 when the difference in ratings between the two players is equal to 200, following the original suggestion by Elo. Moreover, in Elo system the initial rating is not important (the difference of rating between two players are considered) if there are not situations introducing inflation or deflation in the system; this is the case with our dataset, because the group of NBA teams is closed. 
The difference in Elo ratings between the two teams fighting in a match will be the first feature used in the present study: for the initial ratings we will follow \citet{Silver2015} and 1300 will be used; moreover, the classical value of 400 for the logistic parameter will be maintained.

\subsection{The difference in relative victory frequencies}
\label{ss:diff}
A second feature directly quantifying the strength of opposing teams is the difference of their relative victory frequencies. Named \emph{diff} in \citep{Migliorati2020cms}, it can be formally defined as follows: 
\begin{equation}
\label{eq:diff}
diff=\frac{won\_matches_{ht}}{played\_matches_{ht}} - \frac{won\_matches_{at}}{played\_matches_{at}}
\end{equation}
where the subscript \(ht\) means home team, and the subscript \(at\) mean away team.
\\ \(Diff\) statistics ranges from -1 to 1, where value 1 means that the home team is absolutely the strongest between the two teams (it won all played games, with a relative frequency of victories equal to 1, where the away team never won: relative frequency of victories equal to 0) and value -1, viceversa, means that away team actually is the strongest. So, \emph{diff} is a clear and concise way for showing the difference in class between the two teams, providing an analytical definition for a classic rule of thumb often used in naive fan predictions  (the team that won more in the (recent) past is the favorite).

\subsection{The Oliver's Four Factors}
\label{ss:4f}
Basketball milestones of this analytics-based approach are pioneering works \citep{Oliver2004, Kubatko2007}, where famous ``Four Factors to success'' were introduced in trying to understand how basketball teams win games. Four Factors are a set of derived statistics, built on the top of classical \emph{box-score} statistics, considered fundamental for winning a match. They are defined beginning from the concept of \emph{possession}, i.e. the number of times a team gains control of the ball during a match, and used for summarizing the attitude of a team with respect to shooting, turnovers, rebounds and free throws as in Equations \ref{4F:sh},\ref{4F:tu},\ref{4F:re},\ref{4F:fr} (refer to Table \ref{var_acr} for variable meaning):
\begin{enumerate} 
	\item Shooting, measured by effective Field Goals percentage:
		\begin{equation}
		\label{4F:sh}
		eFG\% = (P2M + 1.5*P3M) / (P2A + P3A)
		\end{equation}
	\item Turnover ratio, the number of turnovers (i.e. loss of ball) per possession:
		\begin{equation}
		\label{4F:tu}
		TO\_ratio = TOV / POSS
		\end{equation}
	\item Rebounds, defined by offensive rebounding percentage:
		\begin{equation}
		\label{4F:re}
		OREB\% = OREB / (OREB + DREB)
		\end{equation}
		where, in the denominator, the team offensive rebounds and the opponent team defensive rebounds are considered, respectively.
	\item Free throws rate:
		\begin{equation}
		\label{4F:fr}
		FT\_rate=FTM / (P2A + P3A)
		\end{equation}
\end{enumerate}

\begin{table}[h]
\caption{Variables' acronym meaning}
\label{var_acr}  
\begin{tabular}{ll}
\hline\noalign{\smallskip}
Acronym & Meaning \\
\noalign{\smallskip}\hline\noalign{\smallskip}
P2A & 2-point field goals attempted \\
P3A & 3-point field goals attempted \\
FTA & free throws attempted \\
P2M & 2-point field goals made \\
P3M & 3-point field goals made \\
FTM & free throws made \\
OREB & offensive rebounds \\
DREB & defensive rebounds \\
TOV & turnovers \\
POSS & possessions \\
\noalign{\smallskip}\hline
\end{tabular}
\end{table}

For each match, the Four Factors for both the home team (marked with $ht$ in the following) and the away team ($at$) can be computed, leading in effect to eight factors. In this work the Four Factors have been calculated using the R package {\fontfamily{pcr}\selectfont BasketballAnalyzeR} \citep{Sandri2018,Sandri2020b}.

\section{The dataset}
\label{s:dataset}
The dataset\footnote{Basketball dataset used in this dissertation has been obtained on the basis of play by play data kindly provided by BigDataBall (www.bigdataball.com), a data provider that leverages computer vision technologies to enrich and extend sports data sets with a number of unique metrics: since its establishment, BigDataBall has supported many academic studies as a reliable source of validated and verified statistics for NBA and several other sports.} includes NBA matches from season 2004-2005 to season 2019-2020 (until 11/03/2020, when NBA was stopped for a period due to Covid19). During the 16 seasons taken into account, some franchises have been subject to changes, so the total number of different teams in the dataset should be 34. For our analyses we adopted the most recent names (i.e. Brooklyn Nets, New Orleans Pelicans and Oklahoma City Thunder) for each franchise affected by changes in the considered period, reducing the number of teams in the dataset to the canonical 30.
\\Only regular seasons have been included in the dataset, discarding playoff matches, trying to privilege uniformity of data; usually, each single season is seen as a uniform period where teams are perceived as homogeneous entities. The assumption is that in a single season there is continuity for a team, at least in its fundamental aspects, whereas changes occur between a season and the following one.
\\Actually, the situation seems to be different: it is true that teams can heavily change from one season to another, but several changes also occur during a single season. These changes can impact not only rosters (think to new free agents' contracts, new multi-year contracts, ten-day contracts for facing injuries, player exchanges, ...), but can involve coaches, managers and referees, too. Sources as \citep{Marusek2021, Sports2021} confirm that fact: it is easy to verify how there is a huge number of transactions  not only between seasons, but also during a season, invalidating the perspective of teams as homogeneous entities in that period. During season 2018-19, for instance, there were about 400 off-season signings, but about 300 in-season signings (particularly when playoffs are approaching and the admitted teams need to prepare them).
So, our choice is to include in the dataset only the matches selected on the basis of a homogeneous regulation framework. In sport, rules drive strategies (think to differences addressed by a championship without relegations, as NBA, with respect to a championship with relegation as normally football championships are) and tactics (think to football offside rule, or to NBA zone defense, prohibited until the 2001–2002 season), and it seems fair to consider them in dataset definition.
\\NBA playoff rules are very different from regular season rules, and in the perspective we are sketching (depending of course on the analysis goals) we preferred to avoid including both playoff and regular season games in the same dataset. Instead, despite some existing differences, regular seasons' rules starting from season 2004-2005 are reasonably uniform, and their matches can be fairly included in a single dataset. Of course, the same kind of analysis made in this work about regular seasons could be replicated for playoff games which, in turn, share a not so different frame of rules (in effect, it could be interesting to verify if results found for regular seasons dataset change or not when considering the playoff games dataset. This job is left to future analyses).
\\In the dataset the features based on Elo, \(diff\) and Four Factors have been calculated ex ante, i.e.considering only information from prior matches, to make them suitable for outcome predictions, taking into account:
\begin{itemize}
\item the periodicity, considering both the historical (considering all prior games) and the dynamic perspective (averaging on a subset of prior matches). Moreover, the mechanism of regression to mean \citep{Galton1889} has been implemented for historical features, seeming particularly suitable for NBA \citep{Silver2015}, where at the end of each season there is an attempt to re-balance teams strength.
\item the court where matches have been played (called \emph{the court issue} in the following): besides features usually calculated considering all matches, two new statistics based considering only either home or away data will be calculated, too.
\end{itemize}
The whole dataset initially included 19.138 observations, one for each match. Features calculation introduced some \emph{Not Available} (NA) values to indicate either the absence of a value or the impossibility to calculate\footnote{For instance, when average depth is set to \(n\) and less than \(n\) observations are available. In particular, the first season is affected by this issue.}. Rows containing NA values have been discarded, arriving to a magnitude of about 18.000 observations (depending on values used to calculate features, for instance the depht in the dynamic approach).
\\ The dataset has been classically split in training and testing subsets; the training dataset features' values have been standardized, and also the testing dataset features' values have been modified on the basis of mean and standard deviation used in standardizing corresponding training dataset features.
\\As a summary, Table \ref{tab:feat-kind} reports the ways adopted in features calculation, taking into account both periodicity (historical VS dynamic) and the court issue.
\begin{table}
\small
\caption[Features variants]{Features variants calculated for Elo, \(diff\), Four Factors.}
\label{tab:feat-kind}       
\begin{tabular}{lll}
\hline\noalign{\smallskip}
periodicity & court issue \\
\noalign{\smallskip}\hline\noalign{\smallskip}
historical	&  not considered  \\
historical	& considered  \\
dynamic & not considered  \\
dynamic & considered  \\
\noalign{\smallskip}\hline
\end{tabular}
\end{table}
In the following of the current Section, some details about the features calculation are provided.

\subsection{Periodicity}
\label{ssec:fc.1}

The features used as covariates in models for predicting the outcome of a match have been calculated both from an historical (considering all the prior matches included in the dataset) and dynamic perspective (averaging on a subset of prior matches). Under the so-called historical perspective, if we have to predict the outcome of the game \(g\), a generic feature \(f(t,g)\) for a team \(t\),  with \(t\) ranging from \(1\) to the number of teams \(T\), will be calculated as in Equation \ref{fc.eqh}:
\begin{equation}
\label{fc.eqh}
f_{(t,g)}=\frac{\sum_{i=1}^{g-1} f_{(t,i)}}{g-1}
\end{equation}
Under the dynamic perspective,instead, \(f(t,g)\) will be calculated as in Equation \ref{fc.eqd}:
\begin{equation}
\label{fc.eqd}
f_{(t,g)}=\frac{\sum_{i=g-d}^{g-1} f_{(t,i)}}{d}
\end{equation}
where \(d\) is the depth, i.e. the number of prior games to be considered in calculating the average of \(f\). The best\footnote{Some experiments using the exponential smoothing of some past matches as depth for rolling mean has been implemented, but the fitted models show no improvements on the quality of fit with respect to the approach used in the analysis} value of \(d\)is that producing the model with the highest predictions quality. 

\subsubsection{Historical approach: regression to mean}
\label{sssec:fc.rtm}
\begin{figure}
\includegraphics[width=0.7\textwidth]{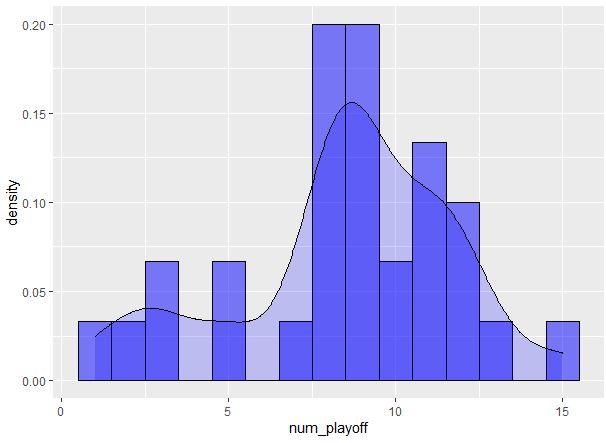}
\caption[Efficiency of NBA re-balancing mechanism]{Efficiency of NBA draft re-balancing mechanism. On the vertical axis the number of playoff accesses bars  and the density function (line) are depicted for seasons from 2004-2005 to 2019-2020.}
\label{fig:fc1}       
\end{figure}

When computing the historical \(f(t,g)\) (equation\ref{fc.eqh}), the regression to mean is used: at each season starting, the features' values are reinitialized, taking into account a  percentage of their past average.
\\Let us consider the last \(N\) regular seasons \(s_1,..,s_N\), each one composed by \(m_k\) matches, where k renages over the number of seasons from 1 to \(N\); moreover, let us consider the generic feature \(f\) of the team \(t\), denoted as  \(f_{t}\), with \(t\) ranging over the number of teams, from \(1\) to \(T\).
\\ The value of a generic feature \(f_{t}\) for the team \(t\) for the first match of the new regular season \(s_{N+1}\), denoted as \(f_{t,s_{N+1}^1}\), is calculated as in \ref{fc.eq1}, adding a proportion 1-P of the value of the feature after the last match of the previous season \(S\), i.e. \(f_{i,s_{N}^{m_N}}\) to a proportion P of the average of the values of the feature calculated considering all past matches for all teams:

	\begin{equation}
	\label{fc.eq1}
	f_{t,s_{N+1}^1}=f_{t,s_{N}^{m_N}}*(1-P)+\frac{\sum_{j=1}^T\sum_{k=1}^N\sum_{z=1}^{m_k} f_{j,s_k^z}}{T*\sum_{z=1}^{N} m_z}*P
	\end{equation}
where \(P\) is the proportion of regression to mean to be considered.
A regression proportion equal to 0 reduces equation \ref{fc.eq1} to equation \ref{fc.eq1.2}
	\begin{equation}
	\label{fc.eq1.2}
	f_{t,s_{N+1}^1}=f_{t,s_{N}^{m_N}}
	\end{equation}
where the first value of the feature for the new season of a team is equal to the last value of the feature of the previous season for that team; this means to have continuity among seasons, without any regression to mean: like a single, long season for the entire period 2004-2020.\\
At the other opposite, when the regression proportion \(P\) is equal to 1, equation \ref{fc.eq1} is reduced to equation \ref{fc.eq1.3}
	\begin{equation}
	\label{fc.eq1.3}
	f_{t,s_{N+1}^1}=\frac{\sum_{j=1}^T\sum_{k=1}^N\sum_{z=1}^{m_k} f_{j,s_k^z}}{T*\sum_{z=1}^{N} m_z}
	\end{equation}
meaning that the starting features' values for each season are the same for every team, equal to the mean of all past values; it is a complete regression to mean for all teams.
\\The mechanism of regression to mean is suitable not only from a statistical point of view \citep{Galton1889}, because extreme values tend to become closer to the mean in new observations, but it seems particularly suitable for NBA \citep{Silver2015}, where the draft mechanism is adopted: at the end of each season, the worst classified teams will have the precedence in selecting new players; at the opposite, the best classified teams will be the last in such a choice. The draft mechanism is not perfect, but ensures a certain balancing among teams, see Figure \ref{fig:fc1} where the number of playoff accesses for each team is used as a measure of the efficiency of draft mechanism. For a good balancing, each team should have the same number of playoff accesses (about eigth in the considered period), with a density curve (depicted in light blue) different from zero only around that value. Instead, there are teams with only one or two accesses, and other teams with 15 accesses. In this work we verified (see Section \ref{sec:ann:res}) how regression to mean is fundamental for having good predictions quality for models using historical features as regressors. 

\subsection{The home factor}
\label{ssec:fc.2}

In NBA, the court factor plays an important role \citep{Harville1994,Jones2007}. The analysis about this topic on the dataset here considered confirms that: on average, home teams win 59.27\% of the matches; a summary per season is reported in Table \ref{tab:ha.tab}, where home victories percentage is always greater than 57\% (apart from the season 2019-2020, for which data are limited to 20/03/20) and in several seasons the 60\% is exceeded.

\begin{table}
\small
\caption[Percentage of home victories per regular season]{Percentage of home victories per regular season (season 2019-2020 is limited to 20/03/20)}
\label{tab:ha.tab}       
\begin{tabular}{lclc}
\hline\noalign{\smallskip}
season & home victories \% & season &  home victories \% \\
\noalign{\smallskip}\hline\noalign{\smallskip}
2004-2005	& 60.43 & 2012-2013	& 61.19 \\
2005-2006	& 60.46 &2013-2014	& 58.05 \\
2006-2007	& 59.15 &2014-2015	& 57.48 \\
2007-2008	& 60.18 &2015-2016	& 58.86 \\
2008-2009	& 60.73 &2016-2017	& 58.37 \\
2009-2010	& 59.40 &2017-2018	& 57.89 \\
2010-2011	& 60.33 &2018-2019	& 59.27 \\
2011-2012	& 58.59 &2019-2020	& 55.10 \\
\noalign{\smallskip}\hline
\end{tabular}
\end{table}


This information should be considered in features' calculation as described by next sections. 

\subsubsection{Modifying Elo calculation to account for the home advantage}
\label{ssec:fc.h.elo}
Elo definition is usually modified to take into account the home court advantage (chess does not care about the home factor, but many other sports, and basketball among them as demonstrated above, should): if home victory is more frequent, the impact on the ratings' update for the home team victory should be decreased. The classic approach to do that consists in adding a penalization parameter to the exponent, as in Formulas \ref{elo.eq1new} and \ref{elo.eq2new}; in this way, a home victory will produce a smaller effect on rating updates, balancing the home court factor:
\begin{equation}\label{elo.eq1new}
P(p1w)=\frac{1}{1+10^{\frac{-(R1-R2+HA)}{400}}}
\end{equation}
\begin{equation}\label{elo.eq2new}
P(p2w)=\frac{1}{1+10^{\frac{-(R2-R1-HA)}{400}}}
\end{equation}
This penalization parameter must be carefully quantified because, as shown in Table \ref{tab:ha_values}, it can play an important role in Elo rating update\footnote{in \citet{Silver2015} home-court advantage is quantified as 100 Elo points}. The Table \ref{tab:ha_values} contains some examples of rating updates for a match involving two teams with same Elo rating (1300) before their match. On the basis of the value of penalization parameter (column Home\_adv), the result of the application of Equations \ref{elo.eq1new} and \ref{elo.eq2new} to calculate the probability of victory for the home team (column\(P(p1w)\)) and for the away team (column \(P(p2w)\)) changes and consequently, on the basis of the result of the match (column result), the Elo rating values after the match (columns newElo(p1) and newElo(p2), respectively) can greatly change, too.
\begin{table}
\small
\caption[Impact of home advantage penalization parameter]{Examples of the impact of home advantage penalization parameter on ratings update for two teams with same Elo rating (1300) before their match. Column Home\_adv contains the home advantage, columns P(p1w) and P(p2w) the probability of victory for home and away team, respectively, Column result contains 1 in case of victory of home team and 0 otherwise, Columns newElo contain the update of Elo ratings for player 1 (p1) and player 2 (p2), respectively}
\label{tab:ha_values}       
\begin{tabular}{llllll}
\hline\noalign{\smallskip}
Home\_adv &  \(P(p1w)\) & \(P(p2w)\) & result & newElo(p1) & newElo(p2)\\
\noalign{\smallskip}\hline\noalign{\smallskip}
0	& 0.50 & 0.50 & 1 & 1315.00 & 1285.00 \\
50 &  0.57 & 0.43 & 1 & 1312.86 & 1287.14\\
100  &  0.64 & 0.36 & 1 & 1310.80 & 1289.20\\
150  & 0.70 & 0.30 & 1 & 1308.90 & 1291.10 \\
\hline
0	& 0.50 & 0.50 & 0 & 1285.00 & 1315.00\\
50 &  0.57 & 0.43 & 0& 1282.86 & 1317.14\\
100  &  0.64 & 0.36 & 0 & 1280.80 & 1319.20\\
150  & 0.70 & 0.30 & 0 &  1278.90 & 1321.10\\
\noalign{\smallskip}\hline
\end{tabular}
\end{table}
\\The Elo formula can be further generalized as in equations \ref{elo.eq1new2} and \ref{elo.eq2new2}: for each team, two parameters \(\alpha_{adv}\) and \(\beta_{dis}\) (prize and penalization, respectively) are added to the numerator of the exponent, and the value assigned to them represents the sums of all advantages and disadvantages for that team:
\begin{equation}\label{elo.eq1new2}
P(p1w)=\frac{1}{1+10^{\frac{-(R1-R2+\alpha_{1_{adv}}-\beta_{1_{dis}})}{400}}}
\end{equation}
\begin{equation}\label{elo.eq2new2}
P(p2w)=\frac{1}{1+10^{\frac{-(R2-R1+\alpha_{2_{adv}}-\beta_{2_{dis}})}{400}}}
\end{equation}
In this way, in the Elo equations it is possible to take into account not only the home court factor, but also other factors, to be properly quantified, potentially offering some additional information: for instance player injuries (as disadvantage for new injuries, or possible advantage when a top player returns to play, see \citealp{Marusek2021,Hopkins2021} for good data sources), logistics (disadvantages due to travels or court altitude \citet{Silver2020a}), number of days among consecutive matches \citet {Manner2016}, referees \citep{Price2009, Deutscher2015}.
\\In this work only home advantage has been considered (using for Elo rating calculation the {\fontfamily{pcr}\selectfont Elo} R package \citep{Heinzen2020}, leaving the management of other information to future works.

\subsubsection{The court issue}
\label{sssec:fc.h.stats}
Typically, the statistics we are taking into account are calculated considering all matches, without reference to the court where matches are played. In effect, the performances of a team can be very different for matches played at home with respect to matches played away. As an example, in Table \ref{tab:ha.tab2} few observations about some Detroit Pistons (DET) matches are reported (starting of season 2004-2005). For each match the home team, the away team and the result (1 means victory for the home team) are specified. Moreover, the victory relative frequency (named \(ratio\)) is calculated, both as usual, considering all matches (DET ratio column), and differentiating these statistics on the basis of the court (column DET h ratio for the ratio calculated considering only the matches played at home, and column DET a ratio for the ratio calculated considering only the matches played away). After ten matches, the value of the ratio statistics calculated considering all past matches is equal to 0.5 (i.e. Detroit wins one game out of two); but looking to the ratio statistics calculated considering home/away data separation, we have some additional information: that team is really strong at home, with a home ratio equal to 0.8 (they won four out of five games played at home), but instead they are weak when playing away, with an away ratio only equal to 0.2 (they won only one game out of the five played away). 
\\This approach seems to be promising, and it has been adopted in this work: besides features usually calculated considering all matches, two new statistics based on the court issue will be calculated, too.

\begin{table}
\small
\caption[Example of features calculation considering the court factor]{Example of features calculation considering the court factor for some Detroit Pistons matches results. For each match the home team, the away team and the result (where 1 means home team victory) are reported. Moreover, the victory relative frequency is reported, both considering all matches (column DET ratio) and building two separate statistics considering only for home and away matches, respectively (columns DET h ratio and DET a ratio)}
\label{tab:ha.tab2}  
\begin{tabular}{llllll}
\hline\noalign{\smallskip}
home team & away team & result & DET ratio & DET h ratio & DET a ratio\\
\noalign{\smallskip}\hline\noalign{\smallskip}
DET	& HOU & 1 & 1.00 & 1.00 & \\
TOR	& DET & 1  & 0.50 & & 0.00 \\
DET	& PHI & 1  & 0.67 & 1.00 & \\
LAC	& DET & 0 & 0.75 & & 0.50 \\
DEN	&DET & 1 & 0.60 & &0.33 \\
UTA	& DET & 1 & 0.50 & & 0.25\\
DET	& MIN & 1 & 0.57 & 1.00 & \\
DET	& IND & 0 & 0.50 & 0.75 & \\
DET	& CHA & 1 & 0.56 & 0.80 & \\
CHA	& DET & 1 & 0.50 & & 0.20\\
\noalign{\smallskip}\hline
\end{tabular}
\end{table}

\section{Methods and Models: Deep Learning}
\label{sec:ann:mm}

\subsection{Building Deep Learning models}
\label{s:5:ss:2}

\begin{figure}
 \includegraphics[width=1\textwidth]{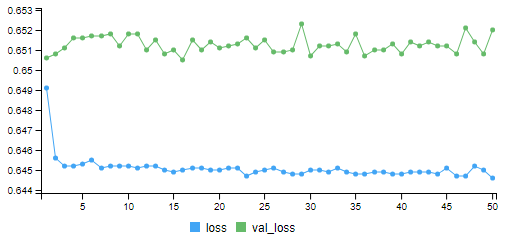}
\caption[Deep Learning model on NBA dataset: overfitting]{Deep Learning model on NBA dataset: example of overfitting (number of epochs on x-axis and measure of loss function on y-axis)}
 \label{fig:k_overfitting} 
\end{figure}

\begin{figure}
 \includegraphics[width=1\textwidth]{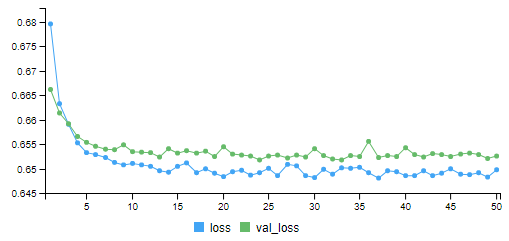}
\caption[Deep Learning model on NBA dataset: reducing overfitting]{Deep Learning model on NBA dataset:  usage of dropout layers and regularization parameters help in reducing overfitting}
 \label{fig:k_no_over} 
\end{figure}

\begin{figure}[h]
 \centering
 \includegraphics[width=0.9\textwidth]{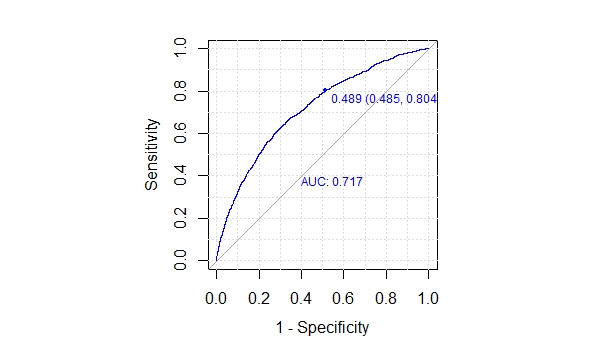}
 \caption[ROC curve for a simple Deep Learning model]{ROC curve for simple (just 1 hidden layers) Deep Learning model on NBA dataset}
 \label{ann:roc_simple}
 \end{figure}

\begin{figure}[h]
   \centering
    \includegraphics[width=0.9\textwidth]{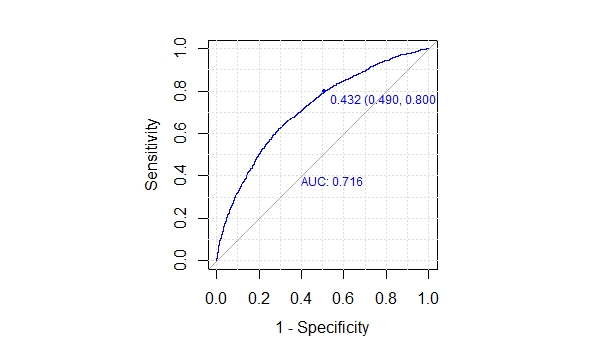}
        \caption[ROC curve for a more complex Deep Learning model]{ROC curve for more complex (ten hidden layers with more units) Deep Learning model on NBA dataset}
       \label{ann:roc_complex}
  \end{figure}

All the models described in this work share the same Deep Learning sequential structure:
\begin{itemize}
\item one first input layer, with a number of input units corresponding to the number of features to be considered in building the model (1 for Elo and \(diff\), 8 for Four Factors (4 for each team))
\item one final output layer, with 1 output unit corresponding to the two possible results of a NBA match
\item a stack of several intermediate hidden sequential layers, connecting the input and output layers. Each hidden layer contains several elaboration units to work on data received from the prior layer before sending them to the following layer.
\\Data transformation on each layer is done by an \(activation  function\): a function, typically non linear, used to transform the weighted sum of layer inputs into outputs; in our models all layers use classic Rectified Linear Activation \(relu\) \citep{Goodfellow2016}, apart from the output layer having a \(sigmoid\) activation function (the most suitable for a two-values classification problem).
\end{itemize}
This traversal process is repeated several times in both directions, with an optimizer updating weight values via a backpropagation algorithm, driven in its action by a loss function to be minimized. In our ANNs:
\begin{itemize}
\item Adam \citep{Kingma2014} is the optimizer.
\item Binary\_crossentropy is the loss function to be minimized.
\item Accuracy is the metric to be used to verify the behavior of the net.
\end{itemize}
As all other machine learning mechanisms, ANN models can be affected by overfitting (i.e. the model is excessively tailored to the training data, and not able to generalize when applied to a different dataset); to verify and tune the behavior of the net before its application to a test dataset, in the  {\fontfamily{pcr}\selectfont Keras} fit function it is possible to reserve a percentage of the training data (excluded from training, in our case 20\%) to validation purposes. Figure \ref{fig:k_overfitting}, with the number of epochs (an epoch consists of one full training cycle on the training set) on \(x\) axis and a measure of the loss function on \(y\) axis, shows an example of overfitting problem, detectable comparing the distance of the loss functions for training (in blue) and for validation (in green) data: the loss function on the training data is decreasing with the increasing of the number of epochs as expected, while the loss function on validation data does not.
\\To reduce overfitting, dropout layers (i.e. hidden layers randomly setting some weight input units to 0, trying to explore not usual paths) and regularization parameters (i.e. penalizations introduced on the weights' matrix to let the network better generalize) can be employed (see Figure \ref{fig:k_no_over}).
The nets, calibrated to produce models with a good prediction quality, are built considering the two hyperparameters (i.e. the number of layers and the number of units for each layer) small in size, a natural consequence of the choice of using a small number of features.
Results show how prediction quality for our classification problem on NBA dataset is almost the same (see Figure \ref{ann:roc_simple} and Figure \ref{ann:roc_complex} ) using  simple nets (with 1 input layer, 1 structural hidden layer and 2 drop layers, 1 output layer; AUC is 0.717, accuracy 0.6721 with a threshold of 0.489) and more complex networks (with ten hidden structural layers and much more units for each layer; AUC is 0.716, accuracy 0.6716  with a threshold of 0.432), consequently the net with the simplest structure has been chosen.

\section{Results}
\label{sec:ann:res}
The results reported in this section have been obtained using a v-fold cross-validation with \(v\)=4 (unless otherwise specified): for each validation, 75\% of observations are randomly selected for training, and 25\% for testing\footnote{A different approach based on time (using the first 14 regular seasons for training and the last 2 regular seasons for testing) was tried, too, producing similar results in terms of prediction quality.}.

\subsection{Using Elo features}
\label{ssec:res1}
Fit quality for Historical Elo based models depends on the three parameters used in calculating the feature: the percentage of regression to mean (see Subsection \ref{sssec:fc.rtm}) and the two values of home advantage (see equations \ref{elo.eq1new} and \ref{elo.eq2new}) and \(K\) (see equations \ref{elo.eq3} and \ref{elo.eq4}) used in Elo rating calculation. To identify which parameters' values produce the best quality, several models have been fitted cycling on possible values of the parameters above, both considering or not the court issue.
\\Predictions quality for models based on dynamic Elo depends on the depth used in averaging, as expressed in equation \ref{fc.eqd}.
Execution results for models based on Elo variants are reported in Table \ref{tab_elo}.
The quality of predictions for models built using historical Elo without considering the court issue is the best one, with an AUC equal to 0.7117 and an accuracy equal to 0,6721 (using a threshold equal to 0.5047). These values have been obtained using a regression to mean percentage \(P\%\) equal to 20, a home advantage parameter equal to 40, and \(K\) equal to 30. 
\\Between the models built using dynamic Elo, the model not considering the court issue, obtained with a depth equal to two, is the best one: its AUC is equal to 0.7117 and its accuracy equal to 0.6736 (using a threshold equal to 0.5049), the best among the models we built in this work. Predictions' quality for the model built using dynamic Elo considering the court issue, obtained with a depth equal to three, has instead an AUC equal to 0.7103 and an accuracy equal to 0.6705 (using a threshold equal to 0.5148).

 \begin{table}
 \small
\caption{\small Best quality of predictions for models based on Elo. For each variant, the best AUC measure, the corresponding threshold and the accuracy measure are reported, together with parameters' values used in Elo calculation}
\label{tab_elo}
\begin{tabular}{llllll}
\hline\noalign{\smallskip}
periodicity & court issue & AUC & threshold & accuracy  &  regression to mean P\% \\
\noalign{\smallskip}\hline\noalign{\smallskip}
historical & not considered & 0.7117 & 0.5047 &  0.6721 & 20 \\
historical & considered	& 0.7001 & 0.5058 & 0.6650 & 60 \\
\noalign{\smallskip}\hline
periodicity & court issue &  AUC & threshold & accuracy & depth \\
\noalign{\smallskip}\hline\noalign{\smallskip}
dynamic & not considered	&  0.7117 & 0.5049 & 0.6736 & 2\\
dynamic & considered	&  0.7103 & 0.5148 & 0.6705 & 3\\
\end{tabular}
\end{table}

\subsubsection{Best accuracy model for Elo}
\label{sssec:res2.3}
\begin{figure}[h]
 \includegraphics[width=0.45\textwidth]{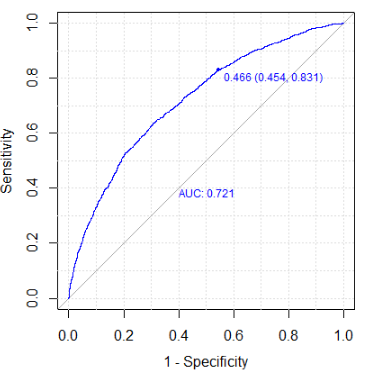}
\caption[AUC for dynamic Elo]{\small AUC for dynamic Elo (single execution). The figure reports the AUC value, together with the optimal threshold 0.466.}
 \label{fig.auc_elo} 
\end{figure}

The best Elo based model uses the dynamic Elo feature (depth equal to 2) without considering the court issue; its AUC, calculated considering a single execution and plotted in Figure \ref{fig.auc_elo}, is equal to 0.7117, the highest among our models.
 Predictions of this model for single seasons have the accuracies reported in Table \ref{tab:seas}: season 2014-2015 shows best accuracy (0.7053), instead worst accuracy (0.6333) is for season 2019/20 (only partially played) and season 2008-2009 (0.6420).


\subsection{Using \(diff\) features}
\label{ssec:res2}

\begin{figure}
 \includegraphics[width=0.7\textwidth]{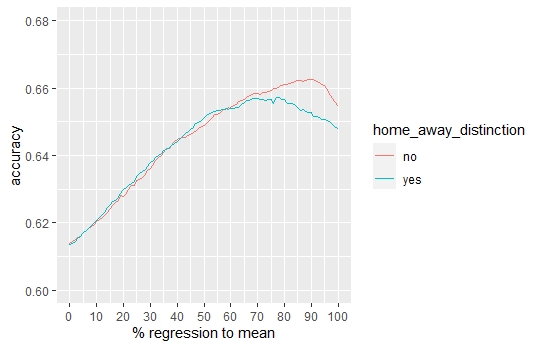}
\caption[Historical diff: accuracy VS regression to mean \%]{\small Historical diff: accuracy VS regression to mean percentage.}
 \label{fig.rm_diff} 
\end{figure}

In historical \(diff\) approach, models' fit quality depends on regression to mean percentage value, as specified in equation \ref{fc.eq1}. As a consequence, in order to identify the model with best predictions quality, all possible values for this parameter have been tried, as reported in figure \ref{fig.rm_diff}, where it is possible to verify how accuracy is really low (about 0.615) when regression to mean is not considered. 

Results reported in Table \ref{tab_diff} show how the quality of predictions of the model built using \(diff\) without considering the court issue is the best one, with an AUC equal to 0.6925 and an accuracy equal to 0.6626 (using a threshold equal to 0.5236). For the model built using dynamic \(diff\), the quality of predictions not considering the court issue is the best one (using a really high depth equal to 50), with an AUC equal to 0.7020 and an accuracy equal to 0,663 (threshold equal to 0.5255).
\begin{table}
\small
\caption{\small Best quality of predictions for models based on \(diff\). For each variant, the best AUC measure, the corresponding threshold and the accuracy measure are reported, together with parameters' values used for calculation}
\label{tab_diff}  
\begin{tabular}{llllll}
\hline\noalign{\smallskip}
periodicity & court issue &  AUC & threshold & accuracy & regression to mean P\% \\
\noalign{\smallskip}\hline\noalign{\smallskip}
historical & not considered	&  0.6925 & 0.5236 & 0.6626 & 90\\
historical & considered	&  0.6775 & 0.4788 & 0.6572 & 78\\
\noalign{\smallskip}\hline
periodicity & court issue &  AUC & threshold & accuracy & depth \\
\noalign{\smallskip}\hline\noalign{\smallskip}
dynamic & not considered	&  0.7020 & 0.5255 & 0.663 & 50\\
dynamic & considered	&  0.6944 & 0.5057 & 0.6586 & 27\\
\noalign{\smallskip}\hline
\end{tabular}
\end{table}

\subsubsection{Best accuracy model for \(diff\)}
\label{sssec:res2.3}
\begin{figure}
 \includegraphics[width=0.5\textwidth]{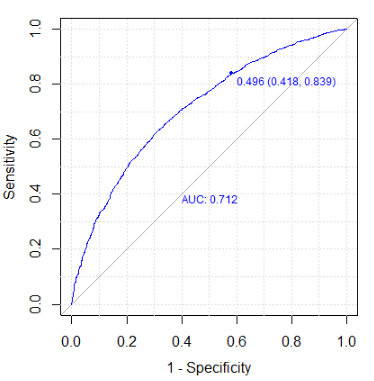}
\caption[AUC for dynamic \(diff\)]{\small AUC for dynamic \(diff\) (single execution). The figure reports the AUC value, together with the threshold (and its coordinates) to be considered in calculating accuracy}
 \label{fig.auc_diff} 
\end{figure}

The best model built on top of \(diff\) feature is that using the dynamic \(diff\) feature without considering the court issue, with a depth in averaging equal to 50; its AUC (calculated considering a single execution) is plotted in Figure \ref{fig.auc_diff}. Predictions for that model considering single seasons have the accuracies reported in Table \ref{tab:seas} (where first season 2004-2005 has been excluded because counting only 35 test observations due to NA omission in dataset preparation). Season 2010-2011 shows best accuracy (0.7034), instead worst accuracy (0.6222) is for season 2019/20 (only partially played) and season 2016-2017 (0.6230).

\subsection{Using Four Factors}
\label{ssec:res3}
\begin{figure}
 \includegraphics[width=0.7\textwidth]{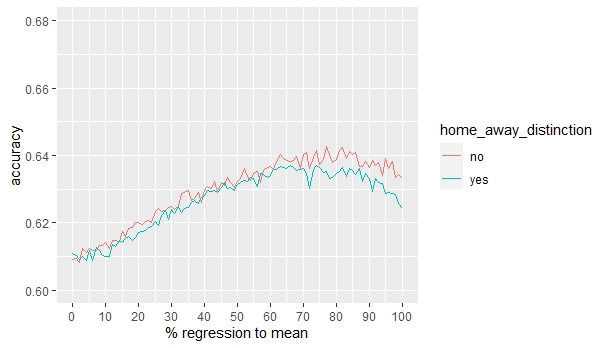}
\caption[Historical Four Factors: accuracy VS regression to mean \%]{\small Historical Four Factors: accuracy VS regression to mean percentage.}
 \label{4f_h_prm}  
\end{figure}

As for other features, predictions quality for historical Four Factors based models depends on the percentage of regression to mean employed in dragging data from one season to the following one; instead, models' quality in predictions depends on the depth employed for calculating rolling mean value, as expressed in equation \ref{fc.eqd}.
Also in this case (as for the corresponding situation for \(diff\)), prediction quality for historical approach without applying regression to mean is actually lower (accuracy measure about 0.61), as reported in Figure \ref{4f_h_prm}. Higher accuracy (0.6427) is found for the model without home/away data separation (considering a regression to mean percentage of 78\%), against 0.6371 for model fitted considering  home/away data separation (with a regression to mean percentage of 74\%).

Table \ref{tab_4f_h} reports some results: the model built on historical Four Factors without considering the court issue is the best one, with an AUC equal to 0.6655 and an accuracy equal to 0.6400 (threshold equal to 0.5334). Between dynamic features, the two  models are equivalent in terms of quality of fit, slightly less than quality of historical model.
\begin{table}
\small
\caption{\small Best quality of predictions for models based on Four Factors. For each variant, the best AUC, the corresponding threshold and the accuracy measure are reported, together with the parameter's value used for calculation}
\label{tab_4f_h}  
\begin{tabular}{llllll}
\hline\noalign{\smallskip}
periodicity & court issue &  AUC & threshold & accuracy & regression to mean P\% \\
\noalign{\smallskip}\hline\noalign{\smallskip}
historical & not considered	&  0.6655 & 0.5334 & 0.6400 & 78\\
historical & considered	&  0.6527 & 0.4968 & 0.6347 & 74\\
\noalign{\smallskip}\hline
court issue &  AUC & threshold & accuracy & depth \% \\
\noalign{\smallskip}\hline\noalign{\smallskip}
dynamic & not considered	&  0.6495 & 0.4934 & 0.6371 & 42\\
dynamic & considered	&  0.6492 & 0.5091 & 0.6372 & 36\\
\noalign{\smallskip}\hline
\end{tabular}
\end{table}

\subsubsection{Best accuracy model for Four Factors}
\label{sssec:res3.3}
\begin{figure}
 \includegraphics[width=0.5\textwidth]{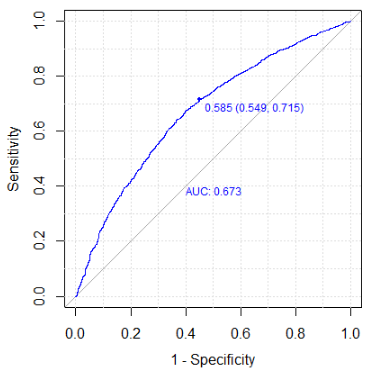}
\caption[AUC for historical Four Factors]{\small AUC for historical Four Factors (single execution). The figure reports the AUC value, together with the optimal threshold 0.585}
 \label{fig.auc_4f} 
\end{figure}

Best predictions are generated using the model based on historical Four Factors without considering the court issue, with a regression to mean percentage P\% of 78; its AUC (calculated considering a single execution) is plotted in Figure \ref{fig.auc_4f}. 

Predictions splitted on seasons (first season, 2004-2005, has been excluded, because counting only 35 observations due to NA management in dataset preparation) produces accuracies reported in Table \ref{tab:seas}, with regular season 2014-2015 showing best accuracy (0.6788) and worst accuracy (0.6175) reported for season 2005/06.


 \begin{table} 
 \small\caption[prediction accuracy per season]{\small prediction accuracy per single season of the best model for each feature; first season 2004-2005 has been excluded because not meaningful for some features}
\label{tab:seas}       
\begin{tabular}{lccc}
\hline\noalign{\smallskip}
season &  best elo accuracy & best \(diff\) accuracy & best Four Factors accuracy\\
\noalign{\smallskip}\hline\noalign{\smallskip}
2005-2006	&   0.6772 &   0.6772 &  0.6175\\
2006-2007	&  0.6885 &  0.6689 &  0.6230\\
2007-2008	& 0.6753 & 0.6786 & 0.6494\\
2008-2009	& 0.6420 & 0.6636 & 0.6235\\
2009-2010	& 0.7028 & 0.6935 & 0.6502\\
2010-2011	&  0.7003 &  0.7034 &  0.6177\\
2011-2012	&  0.6545 &  0.6364 &  0.6455 \\
2012-2013	& 0.6677 & 0.6553 & 0.6522\\
2013-2014	& 0.6829 & 0.6934 & 0.6760\\
2014-2015	& 0.7053 & 0.6623 & 0.6788\\
2015-2016	& 0.6768 & 0.6734 & 0.6566  \\
2016-2017	& 0.6459 & 0.6230 & 0.6590\\
2017-2018	& 0.6633 & 0.6498 & 0.6532 \\
2018-2019	& 0.6964 & 0.6568 & 0.6667\\
2019-2020	& 0.6333 & 0.6222 & 0.6370 \\
\noalign{\smallskip}\hline
\end{tabular}
\end{table}


\section{Conclusions}
\label{sec:ann:con}
In this contribution we showed how appropriately defined statistics can profitably be used as single features in fitting models for outcome predictions on a basketball dataset including 16 NBA regular seasons from 2004-2005 to 2019-2020.
\\The models quality is better than quality of models fitted using Four Factors, a synthesis of \emph{box-score} statistics, and comparable to results reported in the literature (with an accuracy about 0.67\%-70\%).
\\The best prediction quality for a model considering the whole period has been produced using a single dynamic Elo feature (not considering the court issue), with an averaging depth equal to two (i.e. only Elo rating of prior two matches are considered in feature calculation). For this model, the AUC is equal to 0.7117  and the accuracy (using a threshold equal to 0.5049) is equal to 0.6736 (same AUC of the model built using historical Elo, but higher accuracy).  
\\Comparing the accuracy of prediction on single seasons for the three models producing the best results, the dynamic Elo produces the best prediction in 9 seasons, the dynamic \(diff\) in 5 and the Four Factors in 2. The best accuracy for a single season is equal to 0.7053 for the season 2014-2015.
\\In general, quality of models built using \(diff\) based features is close to quality of models built using Elo, and this is an expected result if we take into account how both these features express a direct measure of the strength of a team. Instead, the quality of models based on Four Factors is remarkably the lowest among the three approaches, suggesting how the approaches based on \emph{box-score} statistics can be close to their limit in outcome prediction quality.
\\Results suggest that the court issue approach to features definition, separating data to be used in features calculation on the base of the court, produces predictions comparable in the quality to models based on usual single feature (not considering the court), offering more interpretation details. Regression to mean plays a relevant role in prediction quality, that probably could be improved considering:
\begin{enumerate}
\item a better management of seasons' change in dynamic feature definition. At the beginning of this work, regression to mean was thought just for historical features, supposing only a small number of prior matches have to be considered in dynamic features definition. Instead, results show that in some cases, as reported in Section \ref{sec:ann:res}, the best quality in dynamic models is often obtained considering a not so small depth (50 and 27, 36 and 42). In these cases, it is not difficult to cross two seasons, and regression to mean can play a role. It has been prototypically implemented also for the \(diff\) feature in its dynamic form, and first results confirm how quality of fit is slightly improved by these strategies; this aspect will be deeply investigated in future steps.
\item Analysis and integration of other kinds of information:
\begin{enumerate}
\item injuries, logistics, referees  as proposed by several authors (see Paragraph \ref{ssec:fc.h.elo} about)
\item social networks (as proposed in \citealp{Miller2015}: today sources like Facebook (dated 2004) and Twitter (dated 2006) are old enough to offer information about several past years), breaking news, betting sites, market exchanges 
\item players' characteristics and performances, both as single and with respect to other teammates (see \citealp{Zuccolotto2020} for a review).
\end{enumerate}
\end{enumerate}
Regarding Deep Learning: for this specific classification problem, we obtained good accuracy and limited overfitting maintaining small both the two hyperparameters (levels, units) of the net, heading towards the so called \(shallow\) learning: more complicated nets do not seem to offer great advantages in terms of quality of predictions.
\\Few last words about  {\fontfamily{pcr}\selectfont Keras}, the library we used to build Deep Learning models in R. Our activities have been really simplified by this package, offering a great abstraction on neural nets and enabling to focus just on relevant model aspects. At the moment, there is not a complete default explanation mechanisms associated to that library, but many researches are ongoing to offer explanation facilities (for example see \citealp{Maksymiuk2020}, \citealp{Brandon2020} or \citealp{Molnar2018}), and it is easy to guess how this flaw will be early completely solved.  

\begin{acknowledgement}
I would like to thank Prof. Marica Manisera for the suggestions and the advice she gave me during this work.
\end{acknowledgement}

\bibliographystyle{apalike}
\bibliography{nuovo}
\end{document}